\documentclass[11pt]{article}

\begin{document}
\baselineskip = 0.445cmp
\begin{center}
{\Large {\bf Partial AUC Maximization via \\\vspace{2mm}Nonlinear Scoring Functions}}
\end{center}
\begin{center}
  Naonori Ueda\hspace{1cm} Akinori Fujino\\
  NTT Communication Science Laboratories\\
  2-4 Hikaradai, Seika-cho, ``Keihanna Science City'', Kyoto 619-0237 Japan\\
\vspace*{3mm}

\end{center}

\begin{abstract}
We  propose a method for maximizing a partial area under a receiver operating
characteristic (ROC) curve (pAUC) for binary classification tasks.
In binary classification tasks, accuracy is the most commonly used
as a measure of classifier performance. In some applications such as anomaly detection and diagnostic testing, accuracy is not an appropriate measure since prior probabilties are often greatly biased.
Although in such cases the pAUC has been utilized as a performance measure, few methods have been proposed for directly maximizing the pAUC.
This optimization is achieved by using a scoring function. The conventional
approach utilizes a linear function as the scoring function. In contrast we newly introduce nonlinear scoring functions
for this purpose. Specifically, we present two types of nonlinear
scoring functions based on generative models and deep neural networks. We
show experimentally that 
nonlinear scoring fucntions improve the conventional methods
through the application of a binary classification of real and bogus objects obtained with the Hyper Suprime-Cam on the Subaru telescope.
\end{abstract}

\section{\large Introduction}
The binary classification task, in which the outcomes are
labeled either positive or negative, is one of the most basic problems
in machine learning. Accuracy is the most commonly used performance measure
for the binary classifiers, and it is defined as the proportion of a given set of data that are correctly classified by a classifier.
In some applications such as anomaly detection and diagnostic test, 
prior probabilties are often severely biased. Accuracy is not a
suitable measure when there is an
imbalance between the numbers of positive and negative samples on training samples due to the bias of the priors. 
Instead, we may want to evaluate classifiers not with a simple error
measure with the two types of errors, the {\it false positive rate} (FPR)
and the {\it false negative rate} (FNR). The former (latter) is also
called a type I error (type II error).
In anomaly detection, for example, anomaly data is very
rare compared with normal data and so ususlly pay more attention to
FNR than to FPR.

The area under the receiver operating characteristic
(ROC) curve (AUC) can capture the
relationship between the two types of errors. The ROC curve is
defined by the {\it true positive rate} (TPR=1$-$FNR) and FPR as x and y axes,
respectively, which depicts the relative trade-offs between the
true positive and the false positive when
the classifier's threshold is varied. 
For that reason,  AUC has been used as a
performance measure for binary classifiers, especially under the
condition where the prior probabilities of two classes are biased.  
Several methods for learning a binary classifier by directly maximizing AUC have already been reported \cite{Herschtal}\cite{Rako}\cite{Joachims}\cite{Wang}. 

Recently, however, the partial AUC (pAUC) has attacted
more attension than the AUC because it can focus on a suitable
range of the false positive rate.
For example, in sample imbalanced applications where
there are many fewer positive samples than negative samples,
false positives are intolerable.
Therefore, we wants to increase TPR while keeping FPR
in a range from 0 to a small value. In such a case,
pAUC is more appropriate than AUC as a performance measure.
However, few methods have yet been reported for directly maximizing
pAUC \cite{Narasimhan}\cite{Narasimhan2}.
The pAUC is defined by using a {\it scoring function}, which maps an input sample
to a score (a rank value), and so it is important to design better scoring function for
maximizing pAUC. Previous studies \cite{Narasimhan}\cite{Narasimhan2} proposed a linear function as a scoring function.

In this paper, we present two new types of nonlinear scoring functions based on
generative models and deep neural networks. We show the effectiveness of the proposed
methods through an application involving the binary
classification of real and bogus objects
obtained with the
Hyper Suprime-Cam (HSC) \cite{HSC} on the Subaru telescope.  

\section{\large Partial AUC}
\subsection{\large AUC}
Let $X$ be a sample space. Let ${\cal D}^+$ and ${\cal D}^-$ be
probabilty distributions on $X$. Let $S=(S^+,S^-)$ be a given training sample. Here, $S^+=\{x_1^+,\ldots,x_{n^+}^+\}$ represents $n^+$ positive samples drawn iid according to ${\cal D}^+$, and $S^-=\{x_1^-,\ldots,x_{n^-}^-\}$ represents $n^-$ negative samples drawn iid according to ${\cal D}^-$. Let $f$ be a scoring function:
\begin{equation}
  f:~X\rightarrow \mathcal{R}.
\end{equation}
We can obtain a binary classifier by using $f$ and a threshold $t\in {\mathcal R}$. That is, a sample $x$ is classified as positive (negative) when $f(x)>t$~($f(x)<t$), assuming there is no tie.
Therefore, if $f(x) > f(x')$, then $x$ should be classified as
more positive than $x'$.
The {\it true positive rate} is defined as the probability that the classifier correctly classifies a sample from ${\cal D}^+$ as positive.
\begin{equation}
  \mbox{TPR}_f(t) = P[f(x^+)>t].
\end{equation}
In a similar manner, the {\it false positive rate} is defined as the probability that the classifier misclassifies a sample from ${\cal D}^-$ as positive.
\begin{equation}
  \mbox{FPR}_f(t) = P[f(x^-)>t].
\end{equation}
The {\it receiver operating characteristic curve} (ROC curve) is then drawn by using the values of TPR as
a function of FPT by changing the threshold $t$. Then
the {\it area under the curve} (AUC) is formulated \cite{Dodd} as
\begin{eqnarray}
  \mbox{AUC}&=&P[f(x^+)>f(x^-)]\nonumber\\
    &=& \int_0^1\mbox{TPR}_f(\mbox{FPR}_f^{-1}(u))du.
\end{eqnarray}
Here,
\begin{equation}
  \mbox{FPR}_f^{-1}(u) = \mbox{inf}\{t\in {\mathcal R}\vert\mbox{FPR}_f(t)\leq u\}.
\end{equation}
Given a sample set $S=(S^+,S^-)$, assuming there are no ties, the empirical AUC is given by
\begin{eqnarray}
  \widehat{\mbox{AUC}_f} &=& \frac{1}{n^+n^-}
  \sum_{i=1}^{n^+}\sum_{j=1}^{n^+}I(f(x_i^+)> f(x_j^-))
  \label{auc}.
\end{eqnarray}
\subsection{\large Partial AUC}
The partial AUC~(pAUC) is different from AUC in the sense that
the FPR range is limited when
$\alpha \leq \mbox{FPR}_f(t)\leq \beta$ ($0\leq \alpha<\beta\leq 1$).
Therefore, when $\beta$ is small, pAUC is maximized by  
increasing TPR while keeping small FPR. The pAUC is formulated as 
\begin{equation}
  \mbox{pAUC}(\alpha,\beta)=\frac{1}{\beta-\alpha}
  \int_{\alpha}^{\beta}\mbox{TPR}_f(\mbox{FPR}_f^{-1}(u))du.
\end{equation}
Similarly to AUC,
given $S=(S^+,S^-)$, the empirical partial AUC of $f$ in the FPR range $(\alpha,\beta)$ can be estimated as \cite{Narasimhan}\cite{Narasimhan2}
\begin{eqnarray}
  \widehat{\mbox{pAUC}_f} &=& \frac{1}{n^+n^-(\beta-\alpha)}
  \sum_{i=1}^{n^+}[(j_\alpha - \alpha n^-)
    I(f(x_i^+)> f(x_{(j_\alpha)}^-))
    +\sum_{j=j_{\alpha+1}}^{j_\beta}I(f(x_i^+)> f(x_{(j)}^-))\nonumber\\
    \hspace{2cm}&&+(\beta n^--j_\beta)
           I(f(x_i^+)> f(x_{(j_\beta+1)}^-))]\label{pauc},
\end{eqnarray}
where $j_\alpha=\lceil \alpha n^-\rceil$, $j_\beta=\lfloor \beta n^-\rfloor$,
and $x_{(j)}^-$ denotes the negative instance ranked in $j$th position
among $S^-$ in descending order of scores by $f$. $\lceil a\rceil$ is the smallest integer greater than or equal to $a$, and $\lfloor a\rfloor$ is the largest integer less than or
equal to $a$. $I(z)$ is a Heaviside step function where
$I(z)=1$ if $z$ is true,
$I(z)=0$ otherwise. The derivation of Eq.~(\ref{pauc}) is detailed in \cite{Narasimhan}\cite{Narasimhan2}.
Therefore, the remaining problem is to develop a better scoring
function so that Eq.~(\ref{pauc}) can be larger. 


\section{\large Scoring Functions}
In this section, we discuss the scoring function which plays
an inportant role in maximizing the pAUC.
The scoring function is modeled with a set of
parameters $\theta$ and is therefore denoted by $f(x;\theta)$.
Ideally, the scoring funtion should satisfy
\begin{equation}
  f(x^+;\theta) > f(x^-;\theta)
  \label{sf}
  \end{equation}
for any $x^+\sim{\cal D}^+$ and $x^-\sim{\cal D}^-$.
This enables us to maximize the empirical pAUC given by Eq.~(\ref{pauc})
with respect to $\theta$. However, such ideal scoring function
is difficult to construct, and therefore in practice we try to find a better scoring
function that provides the largest possible empercal pAUC.
Note that since the Heaviside function in Eq.~(\ref{pauc}) is
undifferentiable, it is often approximated by using a logistic
sigmoid function \cite{Yan},
\begin{equation}
  s(x^+,x^-;\theta) = \frac{1}{1+\exp{[-\{f(x^+;\theta)-f(x^-;\theta)\}]}}.
  \label{sigmoid}
\end{equation}
Here $f(x;\theta)$ is also assumed to be differentiable w.r.t. $\theta$.
When the logistic sigmoid function is used, Eq.~(\ref{pauc}) is replaced by
\begin{eqnarray}
  \widehat{\mbox{pAUC}(\theta)} &=& \frac{1}{n^+n^-(\beta-\alpha)}
  \sum_{i=1}^{n^+}[(j_\alpha - \alpha n^-)
    s(x_i^+,x_{(j_\alpha)}^-;\theta)
    +\sum_{j=j_{\alpha+1}}^{j_\beta}s(x_i^+,x_{(j)}^-;\theta)\nonumber\\
    \hspace{2cm}&&+(\beta n^--j_\beta)
           s(x_i^+,x_{(j_\beta+1)}^-;\theta)]\label{pauc2}.
\end{eqnarray}
So, all that remains is to specify the scoring function $f$.
\subsection{\large Linear scoring function}
A linear scoring function has been presented in previous
studies \cite{Narasimhan}\cite{Narasimhan2}; 
\begin{equation}
  f(x;\theta) = \theta^tx.
\end{equation}
Here $a^t$ denotes a transpose of a vector $a$. Although the
linear function is easy to opitimize, it is clear that
a nonlinear function, is more flexible than the linear function and therefore
the former is much more effective than the latter in the sense of Eq.~(\ref{sf})
\subsection{\large Nonlinear scoring function}
\noindent{\bf Deep Neural Networks}\\
In recent years, deep neural networks (DNNs) have been widely
utilized in various fields. The most straightforward way is
to construct a DNN as the nonlinear scoring function $f$. In this setting,
the parameter $\theta$ of a DNN is trained so as to 
maximize the objective given by Eq.~(\ref{pauc2}). In other words,
as shown in the experiment, when training a set of parameters of
a DNN by using $S$, we simply set the objective function as Eq.~(\ref{pauc2}). \\

\noindent{\bf Probabilistic Genarative Models}\\
In another approach, which we newly present in this paper,
we employ generative models. In the machine learning literature,
many researchers have developed and utilized probabilistic generative models
in various applications. 
For example, the Gaussian mixture model has been utilized for real-valued
data. A mixture of multinominals has been
utilized for count data such as bag-of-words for text data. Therefore, it is reasonable that 
the probabilistic generative model will be suitable for
a given application namely designing a scoring function.
This motivated us to consider a scoring
function based on the probabilistic generative models.

Let $p(x;\theta^+)$~($p(x;\theta^-)$) be the
probability distribution of positive (negative) samples\footnote{Note that if $x$ is continuous (discrete), $p$ is a probability density (mass) function.}.
Here, $\theta^+~(\theta^-)$ is the parameter of a probabilistic generative
model for positive (negative) samples.
Then, we consider the following ratio of the probability distribution of positive and negative samples;
\begin{equation}
  r(x;\theta)=\frac{p(x;\theta^+)}{p(x;\theta^-)},
  \label{ratio}
\end{equation}
where $\theta=(\theta^+,\theta^-)$.
Moreover,
let $P(\omega^+)$~($P(\omega^-)$) denote the class prior
of positive (negative) samples. Then the
ratio can be rewritten with the Bayes rule as
\begin{eqnarray}
  r(x;\theta)&=&\frac{p(x;\theta^+)}{p(x;\theta^-)}=
  \left(\frac{P(\omega^+\vert x;\theta^+)p(x;\theta)}{P(\omega^+)}\right)
  \left(\frac{P(\omega^-\vert x;\theta^-)p(x;\theta)}{P(\omega^-)}\right)^{-1}\nonumber\\
  &=& \frac{P(\omega^+\vert x;\theta^+)}{P(\omega^-\vert x;\theta^-)}
  \frac{P(\omega^-)}{P(\omega^-)},
  \label{ratio2}
\end{eqnarray}
where $p(x;\theta)=P(\omega^+)p(x;\theta^+)+P(\omega^-)p(x;\theta^-)$.
Therefore,
\begin{equation}
  \log r(x;\theta)
  = \log P(\omega^+\vert x,\theta^+) - \log P(\omega^-\vert x,\theta^-) +\mbox{const}.
  \label{ratio3}
\end{equation}
Here, the last term of Eq.~(\ref{ratio3}) is a constant independent of $x$.

If $x$ is a positive class sample, then
$\log P(\omega^+\vert x,\theta^+) > \log P(\omega^-\vert x,\theta^-)$ should hold.
Likewise, if $x$ is a negative class sample, then
$\log P(\omega^+\vert x,\theta^+) < \log P(\omega^-\vert x,\theta^-)$ should hold.
This means that the ratio given by Eq.~(\ref{ratio3}) is appropriate as a scoring function.
Therefore, we define the scoring function as
\begin{equation}
  f(x;\theta) = \log \frac{p(x;\theta^+)}{p(x;\theta^-)}
  \label{nlsc}
  \end{equation}
Using Eq.~(\ref{nlsc}), we can employ various kinds of generative models, depending on
application. Typically, a mixture of Gaussian model is useful for continous data, while a
mixture multinomial model is useful for discrete data.
In a conventional generative model approach to classification tasks, each positive and negative model is independently trained by using positive or
negative samples. In contrast, the parameters of
positive and negative models are simultaneously optimized
so that Eq.~(\ref{pauc2}) is maximized. Moreover, in a related matter,
in the context of statistical hypothesis testing, the Neyman-Pearson
lemma states that the likelihood ratio test, in which the likelihood ratio of the two classes is used as the test statistic, is optimal in the sense that it is the most powerful \cite{Neyman}.

\section{Experiments}
\subsection{Test Collection}
We used the 
Hyper Suprime-Cam (HSC)
dataset reported in~\cite{HSC} for an empirical evaluation of
our proposed method. The HSC dataset includes the image
data of 487 real and 267074 bogus optical transient objects
collected with the HSC using
the Subaru telescope. The image data are represented by using thirteen
features as shown in \cite{Morii}, and each image is assigned with
either real (positive class) or bogus (negative class) labels.

We examined the performance of the proposed method in an imbalanced
binary classification task whose aim was to predict the true label of each
image from the feature vectors of the image. In our experiments, we used
70 percent of the image data to train classifiers designed with the
proposed method and the remaining 30 percent to test classifier performance. We formed five different evaluation sets by randomly
dividing the image data into training and test data.
\subsection{Experimental Setting}
We compared the performance of the proposed method with that of conventional
methods, namely DNN-CE, GMM, SVM-pAUC, SVM-AUC, SVM, and RF.

The DNN-CE classifier is a neural network classifier constructed by using
{\em TensorFlow}~\footnote{https://www.tensorflow.org/}, where a
cross-entropy objective function is employed to train the neural
networks. The number of hidden layers was selected from candidate
values of $\{1,2,3,4\}$, and the unit number of each layer was selected
from candidate values of $\{50 n | 1 \le n \le 20, n \in I\}$.  We
used either ``tanh'' or ``selu'' activation functions, and tuned the
weight of L1-regularizer between $10^{-3}$ and $10$.  The dropout ratio
was selected from candidate values of $\{0, 0.2, 0.3, 0.4, 0.5, 0.6,
0.7, 0.8\}$.

The GMM classifier is a generative classifier based on Gaussian mixture
models (GMMs) designed for each class. We trained the classifier by the
maximum likelihood estimation of the GMM parameters. The number of Gaussian
models used for each class was selected from candidate values of
$\{n|1 \le n \le 19, n \in I\}$.

The SVM-pAUC classifier was trained by using ${\rm
SVM}_{pAUC}$-tight~\footnote{http://clweb.csa.iisc.ernet.in/harikrishna/Papers/SVMpAUC-tight/},
which is a pAUC-optimized SVM implementation introduced in
\cite{Narasimhan}\cite{Narasimhan2}.
We examined the performance of SVM-pAUC
classifiers whose $\beta$ values were set at 0.01, 0.05, and 0.1.  We
also examined the performance of an SVM-AUC classifier trained to
maximize the AUC performance for training data. The SVM-AUC classifiers
were obtained by setting $\beta = 1$ in the objective function of the
SVM-pAUC classifier. Moreover, we examined the performance of an SVM
classifier trained by using
{\em sklearn.svm.SVC}~\footnote{http://scikit-learn.org/stable/modules/generated/sklearn.svm.SVC.htm}.
For the SVM-pAUC, SVM-AUC, and SVM classifiers, we used a
linear kernel and tuned their cost parameter values $C$ between
$10^{-3}$ and $10^3$.
The RF classifier is a random forest classifier constructed by using
{\em sklearn.ensemble.RandomForestClassifier}~\footnote{http://scikit-learn.org/stable/modules/generated/sklearn.ensemble.RandomForestClassifier.html}.

We set $100$ for the {\em n\_estimators} option, either ``gini'' or ``entropy''
for the {\em criterion} option, an integer between $1$ to $13$ for the
{\em max\_features} option. We used our
proposed method to train the scoring functions of binary
classifiers designed with a neural network (DNN) and a Gaussian mixture
model (GMM). In this paper, we call the DNN-based classifier DNN-pAUC and
the GMM-based classifier GMM-pAUC.  By using the Adam optimizer included
in {\em TensorFlow}, we estimated the parameter values of the DNN-based
and GMM-based scoring functions that maximize the objective function
shown in Eq.~(\ref{pauc}). The hyperparameter values of the
neural network used for the DNN-pAUC classifier were selected from the
same candidate values as for the DNN-CE classifier.  The number of
Gaussian models for the GMM-pAUC classifier was selected from the same
candidate values as for the GMM classifier.
We set $\alpha = 0$ for both the DNN-pAUC and GMM-pAUC classifiers.  We
examined the performance of the DNN-pAUC and GMM-pAUC classifiers whose
$\beta$ values were set at 0.01, 0.05, and 0.1.  We also examined the
performance of the DNN-based and GMM-based classifiers, called DNN-AUC and
GMM-AUC, respectively, that were trained to maximize the AUC performance for
training data. The DNN-AUC and GMM-AUC classifiers were obtained by
setting $\beta = 1$ in the objective functions of the DNN-pAUC and
GMM-pAUC classifiers, respectively.

We tuned the hyperparameter values of the classifiers with a five-fold
cross-validation of the training data. We selected the best combination of
hyperparameter values from all the candidate values for the GMM-pAUC,
GMM-AUC, GMM, and RF classifiers with a grid search. We tuned the
hyperparameter values of the DNN-pAUC, DNN-AUC, DNN-CE, SVM-pAUC, SVM-AUC,
and SVM classifiers with a tree-structured Parzen estimator, {\em
hyperopt}~\footnote{http://hyperopt.github.io/hyperopt/}, which is a
Bayesian optimization method.

\subsection{Results}

\begin{table}
\begin{center}
\caption{Average pAUC value(\%)} \label{pAUC-result}
{\tabcolsep=3.5pt \small 
\begin{tabular} {c|c|c|c} \hline
Method & $FPR=0.01$ & $FPR=0.05$ & $FPR=0.1$ \\ \hline \hline
GMM-pAUC ($\beta=0.01$) & 31.4 (2.7) & 63.5 (5.1) & 76.0 (3.9) \\
GMM-pAUC ($\beta=0.05$) & 31.6 (0.8) & 67.8 (1.6) & 79.9 (1.7) \\
GMM-pAUC ($\beta=0.1$) & 31.7 (2.0) & 67.1 (2.7) & 79.8 (2.3) \\ \hline
DNN-pAUC ($\beta=0.01$) & 16.1 (5.3) & 43.9 (10.0) & 60.4 (10.4) \\
DNN-pAUC ($\beta=0.05$) & 25.1 (5.3) & 57.6 (9.1) & 71.4 (7.8) \\
DNN-pAUC ($\beta=0.1$) & 24.0 (2.8) & 57.9 (2.9) & 73.0 (2.5) \\ \hline
SVM-pAUC ($\beta=0.01$) & 12.4 (6.5) & 40.6 (10.1) & 55.9 (8.8) \\
SVM-pAUC ($\beta=0.05$) & 13.8 (4.6) & 45.2 (6.0) & 60.9 (4.9) \\
SVM-pAUC ($\beta=0.1$) & 18.9 (1.5) & 48.9 (2.8) & 63.6 (2.3) \\ \hline \hline
GMM-AUC & 29.3 (6.6) & 66.3 (3.5) & 79.6 (1.5) \\
DNN-AUC & 23.0 (4.1) & 56.9 (2.3) & 72.1 (1.7) \\
SVM-AUC & 16.0 (2.1) & 42.9 (3.5) & 58.0 (2.7) \\ \hline \hline
GMM & 22.6 (1.5) & 61.5 (2.3) & 75.5 (1.7) \\
DNN-CE & 13.9 (7.7) & 38.6 (15.7) & 53.4 (17.2) \\
SVM & 4.4 (3.7) & 17.9 (8.9) & 30.4 (13.2) \\
RF & 27.5 (2.7) & 56.4 (4.0) & 67.3 (4.6) \\ \hline \hline
\end{tabular}}
\end{center}
\end{table}

\begin{table}
\begin{center}
\caption{Average TPR values (\%)} \label{TPR-result}
{\tabcolsep=3.5pt \small 
\begin{tabular} {c|c|c|c} \hline
Method & $FPR=0.01$ & $FPR=0.05$ & $FPR=0.1$ \\ \hline \hline
GMM-pAUC ($\beta=0.01$) & 47.4 (3.4) & 83.6 (4.6) & 91.8 (2.8) \\
GMM-pAUC ($\beta=0.05$) & 49.7 (1.0) & 88.4 (2.6) & 94.0 (2.9) \\
GMM-pAUC ($\beta=0.1$) & 50.5 (2.5) & 88.1 (3.0) & 94.7 (1.8) \\ \hline
DNN-pAUC ($\beta=0.01$) & 29.0 (9.3) & 66.8 (14.2) & 84.7 (9.0) \\
DNN-pAUC ($\beta=0.05$) & 43.2 (8.1) & 79.3 (9.0) & 88.9 (4.8) \\
DNN-pAUC ($\beta=0.1$) & 41.9 (4.4) & 81.6 (2.5) & 92.3 (3.0) \\ \hline
SVM-pAUC ($\beta=0.01$) & 23.7 (10.2) & 62.7 (10.5) & 78.8 (5.1) \\
SVM-pAUC ($\beta=0.05$) & 27.1 (8.1) & 68.2 (5.1) & 83.6 (3.6) \\
SVM-pAUC ($\beta=0.1$) & 33.8 (2.8) & 69.2 (2.1) & 85.5 (1.9) \\ \hline \hline
GMM-AUC & 47.1 (7.0) & 89.0 (1.4) & 95.1 (1.9) \\
DNN-AUC & 41.9 (4.7) & 80.3 (1.2) & 92.2 (1.6) \\
SVM-AUC & 28.5 (4.3) & 63.3 (3.8) & 79.9 (1.6) \\ \hline \hline
GMM & 41.4 (3.1) & 85.1 (2.8) & 93.2 (1.4)  \\
DNN-CE & 24.5 (13.2) & 58.8 (20.0) & 75.5 (18.5) \\
SVM & 8.6 (6.5) & 32.6 (15.6) & 54.1 (17.0)  \\
RF & 44.4 (5.6) & 70.3 (4.5) & 82.6 (5.5) \\ \hline
\end{tabular}}
\end{center}
\end{table}

Table~\ref{pAUC-result} show the average pAUC values at three fixed FPR
values, 0.01, 0.05, and 0.1, over the five evaluation sets obtained with
the proposed and compared classifiers. Each number in parentheses in the
table denotes the standard deviation of the pAUC values.
Table~\ref{TPR-result} show the average TPR values at the three fixed
FPR values over the five evaluation sets obtained with the proposed and
compared classifiers. Each number in parentheses in the table denotes
the standard deviation of the TPR values.

As shown in Table~\ref{pAUC-result}, the GMM-pAUC classifiers trained
with $\beta = 0.05$ and $0.1$ provided better average pAUC values than
the GMM-AUC classifier. The DNN-pAUC classifiers trained with
$\beta=0.05$ and $0.1$ also provided better average pAUC values than the
DNN-AUC classifier. These results show that training the GMM- and
DNN-based scoring functions with the proposed method was effective in
improving their pAUC performance.

The performance of the GMM-pAUC classifier trained with $\beta = 0.01$
was worse than that when $\beta = 0.05$ and $0.1$. The performance of the
DNN-pAUC classifier trained with $\beta = 0.01$ was also worse than that
with $\beta = 0.05$ and $0.1$. When the $\beta$ value is small, the number of
negative training samples used to estimate the parameter values of the
GMM-pAUC and DNN-pAUC classifiers is small. Setting too small a $\beta$
value may overfit the classifiers into a small number of negative
training samples.

The GMM-pAUC and DNN-pAUC classifiers outperformed the SVM-pAUC
classifier as shown in Tables~\ref{pAUC-result} and \ref{TPR-result}.
The GMM-pAUC and DNN-pAUC classifiers were designed with non-linear
scoring functions, while a linear kernel was employed for the SVM-pAUC
classifier. These experimental results show that optimizing the non-linear
scoring functions with the proposed method is effective in obtaining
better classifiers for the image classification of real and bogus
optical transient objects.

The GMM-pAUC classifier outperformed the DNN-pAUC classifier, although
neural networks often provide better classifiers. The GMM classifier
provided similar or better performance than the DNN- and SVM-based
classifiers. We assume that the distribution of image data used in
our experiments was better fitted to a GMM-based scoring function, and
thus the GMM-pAUC classifier provided better performance.

\section{Conclusion}
We  have proposed a pAUC maxmization method based on
nonlinear scoring functions for binary classification tasks.
Specifically, we have presented two types of scoring functions;
a deep neural network based scoring function and
a probabilitic generative model based scoring function.
Through the application of the binary classification of
real and bogus objects obtained with the Hyper Suprime-Cam
on the Subaru telescope, we have experimentally comfirmed that 
nonlinear scoring functions outperfom the conventional linear
scoring function for pAUC maximization. 
It is worth mentioning that the results
of the probabilistic generative model based scoring function, 
which is proposed in this paper, were
better than those of a deep neural network. 
\section* {Acknowledgments}
This work is supported by Core Research for Evolutionary Science and Technology (CREST), Japan Science and Technology Agency (JST). We thank the Subaru Hyper Suprime-Cam team for providing a
real data set.



\end{document}